\documentclass{article}

\PassOptionsToPackage{numbers}{natbib}

     \usepackage[preprint]{neurips_2019}

\usepackage{amsmath}
\usepackage[utf8]{inputenc} 
\usepackage[T1]{fontenc}    
\usepackage[bookmarks=false]{hyperref}       
\usepackage{url}            
\usepackage{booktabs}       
\usepackage{amsfonts}       
\usepackage{nicefrac}       
\usepackage{microtype}      
\usepackage{multirow}
\usepackage{multicol}
\usepackage{graphicx,wrapfig}
\usepackage{xcolor}
\newcommand{\abs}[1]{\left|#1\right|}

\title{You May Not Need Order in Time Series Forecasting}

    \author{
  Yunkai Zhang \\
  UC Santa Barbara\\
  \texttt{yunkai$\_$zhang@ucsb.edu} \\
  \And
    Qiao Jiang \\
  Brown University\\
  \texttt{qiao$\_$jiang@brown.edu} \\
  \And 
  Shurui Li \\
  UC Los Angeles\\
  \texttt{shuruili@ucla.edu} \\
  \And 
  Xiaoyong Jin \\
  UC Santa Barbara \\
  \texttt{x$\_$jin@ucsb.edu} \\
  \And 
  Xueying Ma \\
  Columbia University \\
  \texttt{xm2209@columbia.edu}\\
  \And 
  Xifeng Yan \\
  UC Santa Barbara \\
  \texttt{xyan@cs.ucsb.edu}
 }

\begin{document}

\maketitle

\begin{abstract}

Time series forecasting with limited data is a challenging yet critical task. While transformers\citep{attnalluneed} have achieved outstanding performances in time series forecasting, they often require many training samples due the large number of trainable parameters. In this paper, we propose a training technique for transformers that prepares the training windows through random sampling. As input time steps need not to be consecutive, the number of distinct samples increases from linearly to combinatorially many. By breaking the temporal order, this technique also helps transformers to capture dependencies among time steps in finer granularity. We achieve competitive results compared to state-of-the-art on real-world datasets.
\end{abstract}

\section{Introduction}

Time series forecasting is often the key to effective decision making. For example, estimating the demand for taxis over time can help drivers to plan ahead and decrease the wait time for passengers \citep{uber}. For such tasks, the statistical community has developed many well-known forecasting models, such as State Space Models (SSMs) \citep{exposmooth} and Autoregressive (AR) models. However, most of them are designed to fit each time series independently or can only handle a small group of time series instances, while often requiring extensive manual feature engineering \citep{kalmanfilter}. This makes accurate forecasting particularly challenging when we have a large number of time series but each with limited time steps. To address this issue, \citep{genseq, deepar, statespace} proposed to model time series with Recurrent Neural Networks (RNN), such as Long-Short Term Memory Networks (LSTM) \citep{lstm}. One notable drawback is that RNNs force past elements to be stored in memory of a fixed size, indicating that RNNs can struggle to capture long-term dependencies \citep{lstmmemory}. 

\citep{attnalluneed} recently proposed the transformer model which leverages the attention mechanism so that it can access any part of the history regardless of the distance. Transformers have achieved superior performance in the area of natural language processing \citep{attnalluneed} and recently also in the area of time series forecasting \citep{tstransformer}. Despite their outstanding capabilities, transformers often need many training samples due to their large number of trainable parameters. Nevertheless, the performance under limited training samples is still critical. One example is to predict the demand for each product in a warehouse \citep{DeepFactor}. Even with records of daily frequencies, several years of data can only aggregate a few hundred time steps, which is not tolerable if the model requires enough data in order to achieve good performance. 
\begin{figure}[t]
    \centering
    \includegraphics[scale=0.25]{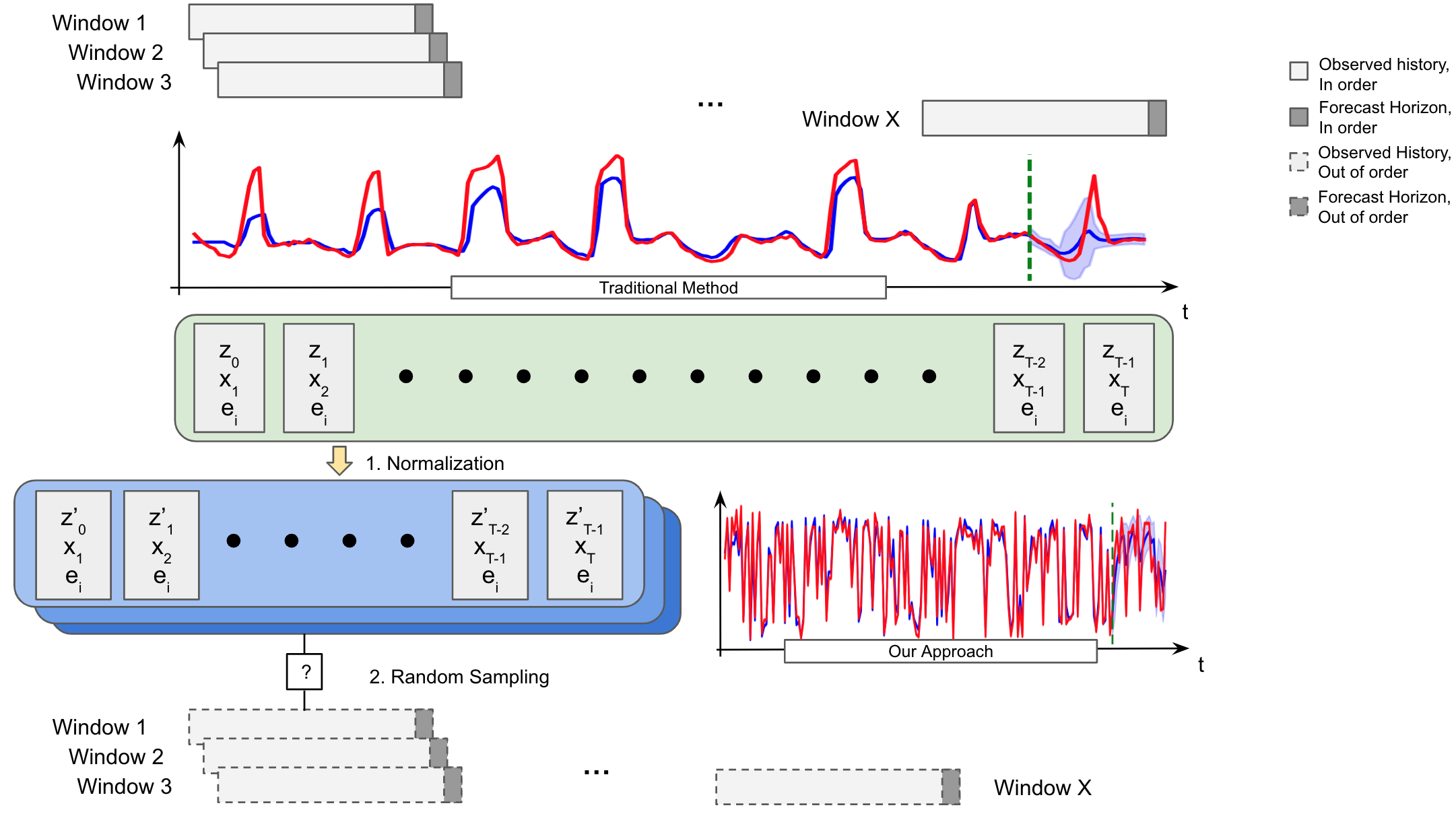}
    \caption{Comparison between the proposed technique and traditional preprocessing. We first normalize the target variables using bigger rolling windows and then randomly sample training windows from the bigger windows or the entire observed history depending on the dataset size.}
    \label{fig:confidence}
\end{figure}
While some research was done focusing on few shot learning for time series forecasting \citep{DeepFactor}, past works on transformers mainly focused on transfer learning in natural language processing through pre-trained input representations \citep{bert, peters2018a, transferNLP}. However, in the domain of temporal point forecasting, it is often more challenging to find a large corpus of similar time series processes since inputs are direct numbers instead of words. On the other hand, some training techniques were also proposed to introduce complexity to the training tasks. BERT \citep{bert} randomly masks some of the input word tokens and predict them in parallel. XLNet \citep{XLNet} suggests to predict all tokens but in random order. RoBERTa \citep{robert} introduces dynamic masking so that different tokens are masked across training epochs.

In this paper, we propose a novel training technique on transformers with applications in time series forecasting. We denoted transformers with this technique as augmented transformers. Through randomly sampling training windows, our contribution is two-fold:
\begin{itemize}
    \item we propose a training technique that expands the number of distinct training tasks from linearly to combinatorially many;
    \item by breaking the temporal order in training windows, augmented transformers can better capture dependencies among time steps.
\end{itemize}

\section{Methodology}

\subsection{Problem Statement}
Given $k$ buildings with conference room utilization levels $[\mathbf{z}^{(i)}_{1:T_i}]_{i=1}^k$ where $\mathbf{z}^{(i)}_t \in [0, 1]$ and a set of associated covariate vectors $[\mathbf{x}^{(i)}_{1:T_i+\tau}]_{i=1}^k$ where $\mathbf{x}^{(i)}_t \in \mathbb{R}^D$, the goal is to predict $\tau$ steps in the future, i.e. $[\mathbf{z}^{(i)}_{T_i+1:T+\tau}]_{i=1}^k$. We denote $\mathbf{z}^{(i)}_{1:t}$ as the observed history and $\tau$ as the forecasting horizon. Formally, we want to model the joint conditional distribution $p([\mathbf{z}^{(i)}_{T_i+1:T_i+\tau}]_{i=1}^k|[\mathbf{z}^{(i)}_{1:T_i}; \mathbf{x}^{(i)}_{1:T_i+\tau}]_{i=1}^k)$. 

We use the autoregressive transformer decoder model from \citep{attnalluneed} by decomposing the joint distribution into the product of one-step ahead distributions $p(\mathbf{z}^{(i)}_{t}|[\mathbf{z}^{(i)}_{1:t-1}; \mathbf{x}^{(i)}_{1:t}]_{i=1}^k)$. The input at each time step is $\mathbf{y}_t = [\mathbf{z}_{t-1}; \mathbf{x}_t; \mathbf{e^{(i)}}]$, where $\mathbf{e^{(i)}} \in \mathbb{R}^E$ is a categorical feature learned from one-hot embedding of each time series instance $i$. Details of the transformer model can be found in Appendix \ref{transformer}.

\subsection{Random Data Sampling}
Traditional data preprocessing for time series forecasting truncate each time series into rolling windows. Some reasons are RNNs often suffer from gradient exploding/vanishing issues \citep{lstmmemory} and transformers need a considerable amount of GPU memory for full attention. Since loss is accumulated over the forecast horizon, training windows should not overlap over the forecast horizon. Each window can be regarded as one training task where the goal is to predict the target values in the forecast horizon given the observed history immediately before the forecast horizon.

Transformers are not aware of relative temporal information as they rely solely on attention mechanism \citep{settrans}. Instead, they resort to either positional sinusoids or learned position embeddings that are added to the per-position input representations \citep{relativepos}. On the other hand, RNNs can model positions relatively through taking input recursively and convolutional neural networks (CNNs) applies kernels based on the relative positions of covered elements \citep{wavenet}. We propose to take full advantage of this unique feature of transformers during model training through effective random data sampling.

As transformers are not aware of relative positions of the time series, they can handle disrupted temporal coherence in the observed history and still able to make reasonable predictions. For example, transformer should yield the same output with $Y_t = [\mathbf{y}_1; \mathbf{y}_2; \cdots; \mathbf{y}_t]$ as $Y_t$ under any random permutations, such as $\hat{Y}_t = [\mathbf{y}_3; \mathbf{y}_{t-5}; \cdots; \mathbf{y}_6]$. 

Thus, for transformers the observed history does not have to come from immediately before. Instead, we can randomly sample all the time steps in each training window independently from the observed history such that they no longer have to be in consecutive temporal order. This technique boosts the number of training tasks from linearly to combinatorially many compared to traditional sampling. 

\subsection{Capture Better Dependencies}
In addition to data augmentation, we argue that the proposed technique also allows the model to extract more complex relationships among time steps. We hypothesize that there are two major reasons. First, it prevents transformers from relying the prediction entirely on one or a few time steps, as these points might not be sampled during training. Second, augmented transformers can capture seasonalities longer than the size of training windows more easily through randomly sampling points beyond the window size. Detailed results are presented in Section \ref{exp}.
\begin{wrapfigure}{r}{6.5cm}
    \includegraphics[scale=0.45]{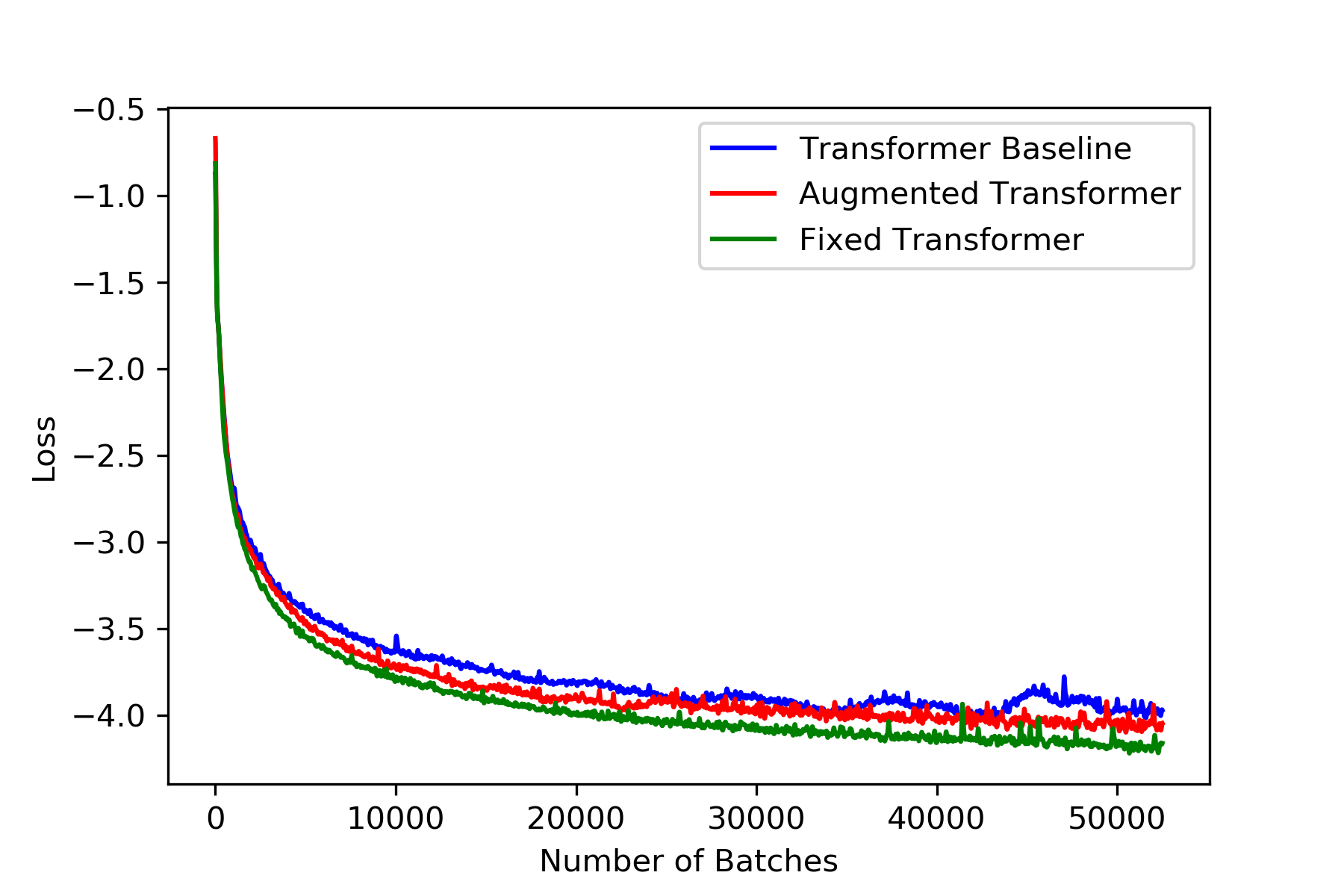}
    \caption{The proposed model converges faster than the vanilla transformer with (augmented) and without (fixed) data augmentation, indicating that it can squeeze out more data dependencies through random sampling.}
    \label{fig:confidence}
\end{wrapfigure} 

Note that inputs to the model need to be normalized. During experiments, we observe that scaling in bigger rollings windows before sampling yields much better performance than scaling after sampling. One reason is that the magnitudes of adjacent time steps are more similar and the relative magnitude in the adjacent window is more important than absolute magnitudes.

Note that some work introduce relative position to transformers, such as through convolutional self-attention \citep{tstransformer} or relative position representations \citep{relativepos}. They are not mutually exclusive with the proposed technique. For example, convolutional attention can be implemented by feeding adjacent time steps as covariates. Further empirical studies are needed.

\section{Experiments} \label{exp}
We conducted experiments on two public benchmark datasets \textit{electricity} \footnote{\url{ https://archive.ics.uci.edu/ml/datasets/ElectricityLoadDiagrams20112014}} and \textit{traffic} \footnote{\url{http://archive.ics.uci.edu/ml/datasets/PEMS-SF}}. The electricity dataset contains the hourly electricity consumption of 370 households from 2011 to 2014. The traffic dataset records hourly occupancy rates (between 0 and 1) of 963 car lanes in San Francisco Bay Area. Following \cite{trmf, statespace}, we use one week of test data for electricity (starting at 12 AM on September 1, 2014) and traffic (starting at 5 PM on June 15, 2008). 

\paragraph{Data Efficiency} We measure the performance of the augmented transformer on the long-term forecasting task presented in \citep{statespace} by directly predicting one week given $\{2, 3, 4\}$ weeks of training data. We compare the augmented transformer against traditional statistical methods as well as recent state-of-the-art deep learning models:
\begin{itemize}
    \item ARIMA: implemented with \texttt{auto.arima} method in R's \texttt{forecast} package;
    \item Exponential smoothing (ETS): implemented with \texttt{ets} method in R's \texttt{forecast} package;
    \item DeepAR\cite{deepar}: an RNN-based autoregressive model;
    \item Deep State Space (DeepSSM)\citep{statespace}: an RNN-based state space model.
\end{itemize}

Following \citep{statespace, DeepFactor}, we use $\rho$-quantile loss to evaluate the prediction accuracy, which is defined as:
\[
    QL_\rho(z, \hat{z}) = \frac{2\sum_{i,t} P_\rho (z_t^{(i)}, \hat{z}_t^{(i)})}{\sum_{i,t}\abs{z_t^{(i)}}},
    P_\rho (z, \hat{z}) =  \begin{cases}
        \rho(z-\hat{z}), & \text{if } z > \hat{z},\\
        (1-\rho)(\hat{z}-z), & \text{otherwise,}
        \end{cases}
\]
where $\hat{z}$ is the empirical $\rho$-quantile of the predicted distribution. $\rho_{0.5}$ and $\rho_{0.9}$ for each model are summarized in Table \ref{tab:long}. Detailed experiment setup can be found in Appendix \ref{experiment}. Overall, augmented transformer surpassed other models in all but one task. More detailed confidence intervals are shown in Appendix \ref{resultdet}. Note that we do not include \citep{DeepFactor} for comparison as we do not limit the number of learnable parameters.

\begin{table*}[ht]
    \centering
    \resizebox{\textwidth}{!}{%
    \begin{tabular}{|c|c|cc|cc|cc|cc||cc|}
    \hline
     & & \multicolumn{2}{c|}{DeepAR} & \multicolumn{2}{c|}{DeepSSM} & 
     \multicolumn{2}{c|}{ARIMA} & \multicolumn{2}{c||}{ETS} & \multicolumn{2}{c|}{Ours} \\
     \hline
    Dataset & Given & $\rho_{0.5}$ & $\rho_{0.9}$ & $\rho_{0.5}$ & $\rho_{0.9}$  & $\rho_{0.5}$ & $\rho_{0.9}$ & $\rho_{0.5}$ & $\rho_{0.9}$ & $\rho_{0.5}$ & $\rho_{0.9}$\\
    \hline
    \multirow{3}{*}{Electricity} & 2 weeks & 0.153 & 0.147 & 0.087 & 0.05 & 0.283 & 0.109 & 0.121 & 0.101 & \textbf{0.083} & \textbf{0.044} \\
     & 3 weeks & 0.147 & 0.132 & 0.130 & 0.110 &0.291 & 0.112 & 0.130 & 0.110 & \textbf{0.083} & \textbf{0.042} \\
     & 4 weeks & 0.125 & 0.080 & 0.130 & 0.110 &0.30 & 0.110 & 0.13 & 0.11 & \textbf{0.084} & \textbf{0.041}\\
    \hline
    \multirow{3}{*}{Traffic} & 2 weeks & 0.177 & 0.153 & 0.168 & 0.117 & 0.492 & 0.280 & 0.621 & 0.650 & \textbf{0.141} & \textbf{0.099}\\
     & 3 weeks & \textbf{0.126} & \textbf{0.096} & 0.170 & 0.113 & 0.492 & 0.509 & 0.529 & 0.163 & 0.140 & 
    0.101\\
     & 4 weeks & 0.219 & 0.138 & 0.168 & 0.114 & 0.501 & 0.298 & 0.532 & 0.60 & \textbf{0.140} & \textbf{0.104}\\
    \hline
    \end{tabular}
    }
    \caption{Evaluation of long-term forecasting (7 days) on electricity and traffic datasets with increasing
training range. For DeepAR and DeepSSM, we use the results reported in \cite{statespace}. The numbers for our model are the averages of 11 trials. Confidence intervals are shown in Figure \ref{fig:confidence}.}
    \label{tab:long}
\end{table*}

\paragraph{Temporal Dependencies} Next, we demonstrate that augmented transformer not only benefits from more distinct training windows, but can also capture temporal dependencies in finer granularity than the original transformer. In this experiment, we limit the number of training windows to be the same as through simple rolling windows to examine the new model without data augmentation. We denote this model without data augmentation as the fixed transformer. We also include \textbf{the vanilla transformer} without random sampling for comparison. The results are shown in Figure \ref{fig:confidence}. For fair comparison, for the vanilla transformer we padded each time series instance with additional zeros so that the rolling windows can start before the beginning of the instance.

The performance of the fixed transformer surpassed the vanilla transformer by a wide margin and is very close to the augmented transformer. This implies that even without data augmentation, random sampling from a larger window during training can help the model extract more features.
\begin{figure}[h]
    \centering
    \includegraphics[scale=0.3]{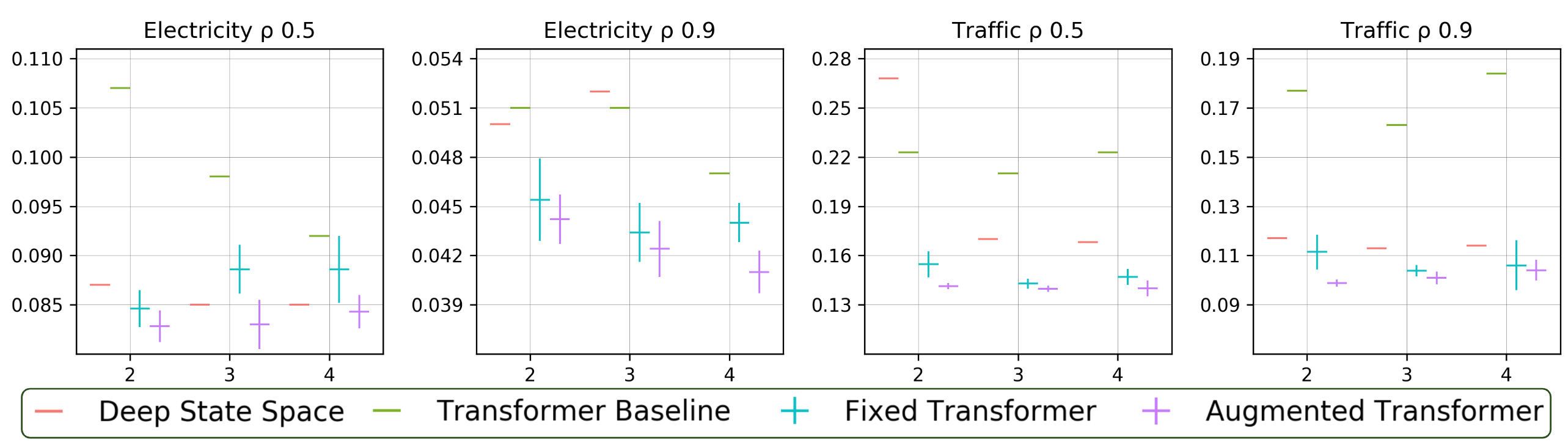}
    \caption{Evaluation of the augmented transformer against strong baselines. Vertical lines indicate confidence intervals generated from 11 trials for augmented transformer and 5 trials for fixed transformer.}
    \label{fig:confidence}
\end{figure}

\section{Conclusion}
We present a novel augmentation technique for transformers on time series forecasting by random sampling of the training windows. The proposed strategy is able to achieve competitive performance compared to strong baselines on real-world datasets. In addition to data augmentation, we also show that augmented transformer can better capture dependencies among time steps.

\section*{References}
\medskip
\small
\renewcommand{\bibsection}{}
\nocite{*}
{\small
\bibliography{main}}
\bibliographystyle{unsrt}

\newpage
\section{Appendix}

\subsection{Transformer} \label{transformer}
The model consists of stacked decoder blocks, where each block has a self-attention layer followed by a feedforward layer. The self-attention layer first transforms $Y_t = [\mathbf{y}_1; \mathbf{y}_2; \cdots; \mathbf{y}_t]$ into $H$ set of attention heads. Let $W_h^Q, W_h^K \in \mathbb{R}^{(1+D+E)\times d_k}$ and $W_h^V \in \mathbb{R}^{(1+D+E)\times d_v}$ be learnable parameters, where $h = 1, \cdots, H$. Each attention head transforms $Y_t$ into query matrices $Q_{h,t} = Y_tW_h^Q$, key matrices $K_{h,t} = Y_tW_h^K$, and value matrices $V_{h,t} = Y_tW_h^V$. From each head, the scaled dot-product attention computes a vector
output for every time step:
$$\mathbf{O}_{h,t}=\operatorname{Attention}\left(\mathbf{Q}_{h,t}, \mathbf{K}_{h,t}, \mathbf{V}_{h,t}\right)=\operatorname{softmax}\left(\frac{\mathbf{Q}_{h,t} \mathbf{K}_{h,t}^{T}}{\sqrt{d_{k}}} \cdot \mathbf{M}\right) \mathbf{V}_{h,t},$$

where $M$ is an upper-triangular mask that prevents the current time step from accessing future information. The feedforward layer then takes the concatenated output from all attention heads and performs two layers of point-wise dense layers with a ReLu activation in the middle. \footnote{Note that we included $M$ in our implementation of augmented transformer, which might not be necessary as the input order is permuted.} Additional details can be found in \citep{attnalluneed}.

\subsection{Result Details} \label{resultdet}

\begin{table*}[ht]
    \centering
    \resizebox{\textwidth}{!}{%
    \begin{tabular}{|c|c|cc|cc||cc|}
    \hline
     & & \multicolumn{2}{c|}{Transformer} & \multicolumn{2}{c||}{Fixed Transformer} & \multicolumn{2}{c|}{Ours} \\
     \hline
    Dataset & Given &  $R_{0.5}$ & $R_{0.9}$ & $R_{0.5}$ & $R_{0.9}$ & $R_{0.5}$ & $R_{0.9}$\\
    \hline
    \multirow{3}{*}{Electricity} & 2 weeks & 0.107 & 0.051 & $0.0846 \pm 0.0019$ & $0.0454 \pm 0.0025$ & $\textbf{0.0828} \pm 0.0016 $ & $\textbf{0.0442} \pm 0.0015 $ \\
     & 3 weeks & 0.098 & 0.051 & $0.0886 \pm 0.0025 $ & $0.0434 \pm 0.0018 $ & $\textbf{0.0830} \pm 0.0025 $ & $\textbf{0.0424} \pm 0.0017 $ \\
     & 4 weeks  & 0.092 & 0.047 & $0.0886 \pm 0.0034 $ & $0.0440 \pm 0.0012 $ & $\textbf{0.0843} \pm 0.0017 $ & $\textbf{0.0410} \pm 0.0013 $\\
    \hline
    \multirow{3}{*}{Traffic} & 2 weeks & 0.223 & 0.177 & $0.1546 \pm 0.0079 $ & $0.1114 \pm 0.0007 $ & $\textbf{0.1413} \pm 0.0019 $ & $\textbf{0.0988} \pm 0.0014 $\\
     & 3 weeks & 0.210 & 0.163 & $0.1428 \pm 0.0031 $ & $0.1038 \pm 0.0023 $ & $\textbf{0.1396} \pm 0.0019 $ & $\textbf{0.1009} \pm 0.0026 $\\
     & 4 weeks  & 0.223 & 0.184 & $0.147 \pm 0.0067 $ & $0.106 \pm 0.0101 $ & $\textbf{0.140} \pm 0.0048 $ & $\textbf{0.104} \pm 0.0042 $\\
    \hline
    \end{tabular}
    }
    \caption{Evaluation of long-term forecasting (7 days) on electricity and traffic datasets with increasing
training range. The confidence intervals shown here are computed from 5 trials for fixed transformer and 11 trials for augmented transformer.}
    \label{tab:long2}
\end{table*}

\subsection{Experiment Setup} \label{experiment}
For each task, our time-based covariate vectors are hour of the day and day of the week. We do not tune hyperparameters heavily. All of the models (both transformer baseline and augmented transformer) use 8 attention heads and dropout of $0.1$. A simple grid-search is used to find the other hyperparamters: $d_k = d_v = \{10, 20\}$,  dimension of the feed-forward layer in the transformer decoder block among $\{20, 30, 40, 50, 60\}$, and embedding dimension of one-hot features among $\{5, 10, 20\}$. 

We scale the time steps with the adjacent $192$ time steps and randomly sample from the entire observed history. For large datasets, sampling from the entire observed history might lead to slower or no convergence. Thus, sampling smaller training windows from bigger rolling windows are recommended. Each training window is of size $192$, where the last $24$ is the forecast horizon. We do not sinusoidal positional embeddings and simply use direct time covariates for faster convergence. With positional embeddings the network achieves around the same accuracy.

Note that during inference, we use all the training data as observed history for the tasks presented as the observed history is limited. For validation set, we randomly sample $10\%$ from the data before the forecast start time and use the rest as the training set. All models are trained on GTX 1080 Ti GPUs.

\end{document}